\documentclass[10pt,twocolumn,letterpaper]{article}

\usepackage[pagenumbers]{cvpr} 

\usepackage{pifont}

\usepackage{multirow}
\usepackage{float}

\usepackage[dvipsnames]{xcolor}

\definecolor{cvprblue}{rgb}{0.21,0.49,0.74}
\usepackage[pagebackref,breaklinks,color links,citecolor=cvprblue]{hyperref}
\newcommand{\spara}[1]{\smallskip\noindent\textbf{#1}}

\def\paperID{} 
\def\confName{}
\def\confYear{}

\title{Grounding-Prompter: Prompting LLM with Multimodal Information for Temporal Sentence Grounding in Long Videos}

\author{Houlun Chen\\
Tsinghua University\\
\and
Xin Wang\thanks{Corresponding author.}\\
Tsinghua University\\
\and
Hong Chen\\
Tsinghua University\\
\and
Zihan Song\\
Tsinghua University\\
\and
Jia Jia$^\ast$\\
Tsinghua University\\
\and
Wenwu Zhu$^\ast$\\
Tsinghua University\\
}

\begin{document}
\maketitle
\begin{abstract}

Temporal Sentence Grounding (TSG), which aims to localize moments from videos based on the given natural language queries, has attracted widespread attention.
Existing works are mainly designed for short videos, failing to handle TSG in long videos, which poses two challenges: i) complicated contexts in long videos require temporal reasoning over longer moment sequences, and ii) multiple modalities including textual speech with rich information require special designs for content understanding in long videos.
To tackle these challenges, in this work we propose a Grounding-Prompter method, which is capable of conducting TSG in long videos through prompting LLM with multimodal information. In detail, we first transform the TSG task and its multimodal inputs including speech and visual, into compressed task textualization. Furthermore, to enhance temporal reasoning under complicated contexts, a Boundary-Perceptive Prompting strategy is proposed, which contains three folds: i) we design a novel multiscale denoising Chain-of-Thought (CoT) to combine global and local semantics with noise filtering step by step, ii) we set up validity principles capable of constraining LLM to generate reasonable predictions following specific formats, and iii) we introduce one-shot In-Context-Learning (ICL) to boost reasoning through imitation, enhancing LLM in TSG task understanding.
Experiments demonstrate the state-of-the-art performance of our Grounding-Prompter method, revealing the benefits of prompting LLM with multimodal information for
TSG in long videos.

\end{abstract}    
\section{Introduction}
\label{sec:intro}

\begin{figure}[t]
  \centering
   \includegraphics[width=\linewidth]{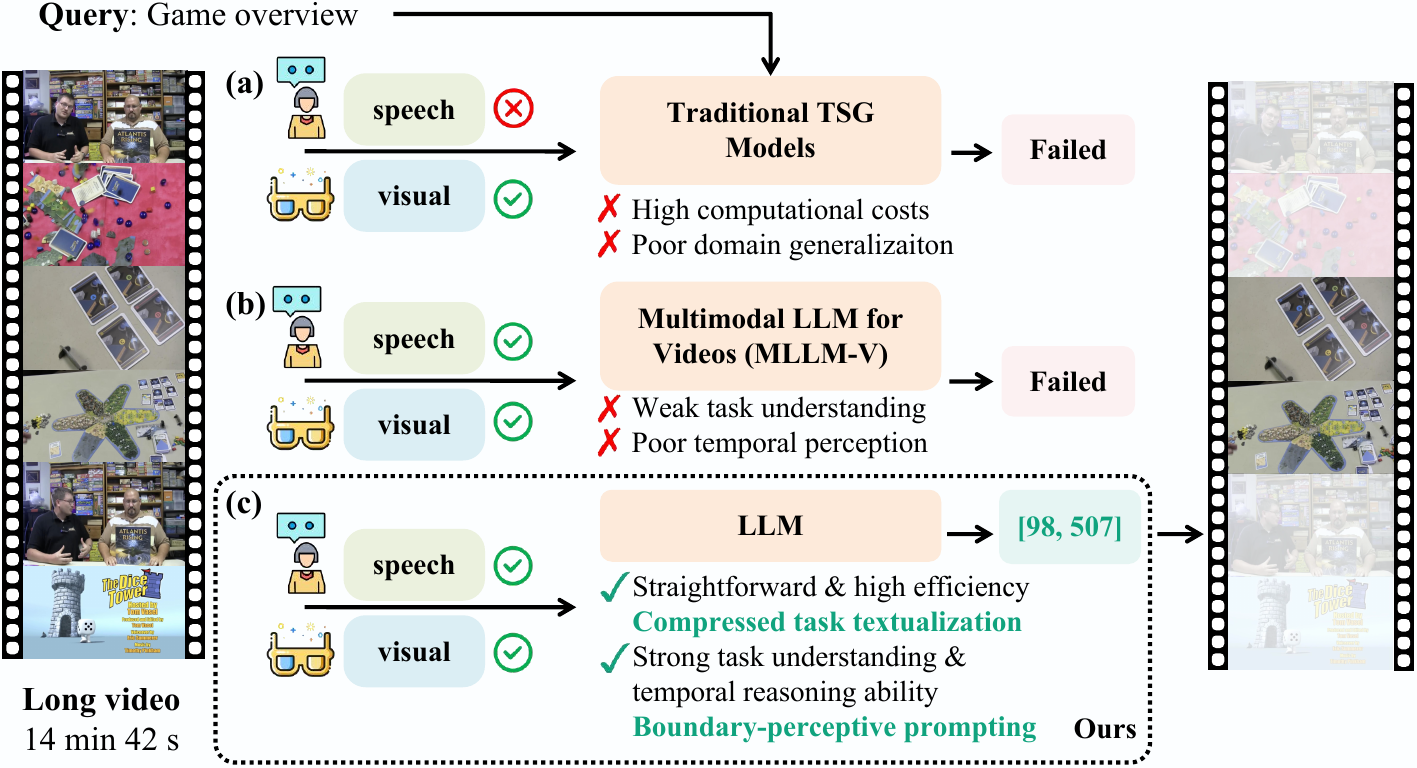}
   \caption{Comparison on technical roadmaps on TSG task.}
   \label{fig:intro}
\end{figure}

Temporal Sentence Grounding~(TSG)~\cite{anne2017localizing,gao2017tall} aims to localize a moment from an untrimmed video to match the given query, requiring methods to understand the temporal boundaries and contexts across videos and texts.

Nevertheless, existing literature~\cite{cao2021pursuit,gao2017tall,yuan2019semantic,duan2018weakly,lin2020weakly,lv2023counterfactual,lin2023univtg,lei2021detecting} are mainly designed for short videos, failing to handle TSG in long videos which is very prevalent in practical scenarios covering movies, news, and courses etc.
However, exploring TSG in long videos encounters the following challenges: 
\begin{itemize}
    \item Long videos normally contain complicated contexts, which require temporal reasoning over longer moment sequences.
    \item Long videos such as movies may contain multiple modalities including textual speech with rich information, requiring special designs for content understanding.
\end{itemize}

On the one hand, it's difficult to extend traditional TSG methods~\cite{cao2021pursuit,gao2017tall,yuan2019semantic,duan2018weakly,lin2020weakly,lv2023counterfactual,lin2023univtg,lei2021detecting} to long videos, as shown in Figure~\ref{fig:intro}(a). With huge amounts of parameters required, they involve high computational costs when fully trained on long-video datasets. Besides, traditional TSG methods suffer from fitting bias of specific datasets~\cite{lan2023survey,yuan2021closer}, thereby exhibiting poor generalization to long moment sequences. Additionally, the incapability of capturing rich semantics from textual speeches prevent them from conducting TSG on long speech-intensive videos. On the other hand, recent trials on employing multimodal large language models for videos~(MLLM-V)~\cite{chen2023videollm,zhang2023video,li2023videochat,maaz2023video} have sprung up with remarkable performance gain across several video tasks. However, these LLM based approaches fail to align the TSG task well with LLMs and thus show poor temporal reasoning ability, especially in long videos, as shown in Figure~\ref{fig:intro}(b).

To handle the challenges above, we reformulate TSG into a long-textual task and propose to empower large language models~(LLMs) with the ability to conduct temporal reasoning under complicated context in long videos. 
Concretely, we propose a novel Grounding-Prompter method via prompting LLMs with speech and visual information to solve the TSG task, as shown in Figure~\ref{fig:intro}(c). First, to align LLMs with the TSG task, we transcribe speeches and caption the sparsely sampled frames that align speeches and scenes with temporal information in order to obtain compressed task textualization. Additionally, to enhance temporal reasoning, we propose a Boundary-Perceptive Prompting strategy, which consists of i) a multiscale denoising Chain-of-Thought~(CoT) that combines global and local semantics with noise filtering step by step, ii) validity principles that constrain LLMs to generate reasonable predictions following specific formats, and iii) one-shot In-Context-Learning~(ICL) that enhances LLMs in TSG understanding and temporal reasoning through imitation.

To verify the superiority of the proposed Grounding-Prompter, we establish a VidChapters-mini dataset for experiments from VidChapters-7M~\cite{yang2023vidchapters}. Empirical results show that our Grounding-Prompter strategy achieves the state-of-the-art performance with great margins compared to other baseline methods. Ablation studies further validate the effectiveness of our designs, demonstrating that our Grounding-Prompter benefits from both textual speech and visual modalities when handling complicated and noisy long contexts around 10k tokens.

To conclude, our contributions lie in the following folds:
\begin{enumerate}
    \item We propose Grounding-Prompter, the first trial to address TSG in long videos through LLM, to the best of our knowledge. 
    \item We integrate textual speech and visual modalities to LLMs with compressed task textualization to handle TSG in long videos, where each modality significantly benefits the predictions. 
    \item We propose a novel Boundary-Perceptive Prompting strategy which enables LLMs to conduct temporal reason over time boundaries correctly under complicated and noisy long contexts around 10k tokens.
    \item We establish a VidChapters-mini dataset and conduct extensive experiments to demonstrate the advantages of Grounding-Prompter over existing baseline methods.
\end{enumerate}

\begin{figure*}[htbp]
  \centering
   \includegraphics[width=\linewidth]{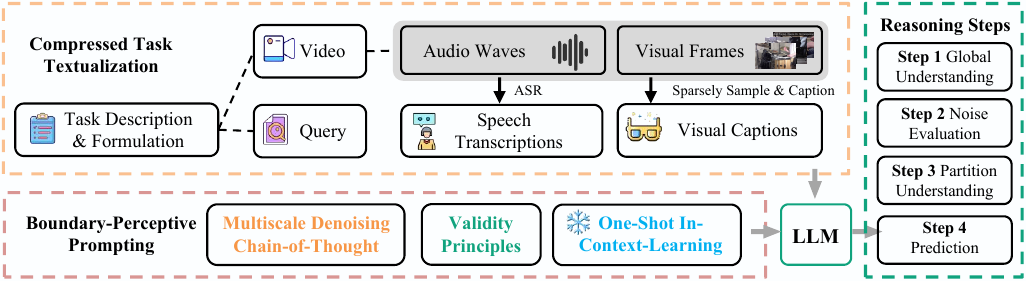}
   \caption{Framework of our Grounding-Prompter. The task inputs are transformed into a compressed textualized representation to feed LLM. To enhance the temporal perception capability, a boundary-perceptive prompting strategy is proposed. Then, the task inputs are rephrased into a fluent prompt under this prompting strategy. LLM is activated to regulate its answer into JSON format and the predictions can be parsed automatically.}
   \label{fig:framework}
\end{figure*}

\section{Related Works}
\label{related}

\spara{Temporal Sentence Grounding~(TSG).}
Temporal Sentence Grounding~\cite{anne2017localizing,gao2017tall} aims to localize a moment from an untrimmed video that matches the given query. Most early TSG literature~\cite{cao2021pursuit,gao2017tall,yuan2019semantic,zhang2019man,zhang2020learning,zhu2023rethinking,li2022end,mun2020local,yuan2019find,zeng2020dense} solves it in a supervised manner. Since they heavily rely on labor-intensive manual annotations, a few works~\cite{chen2020look,gao2019wslln,mithun2019weakly,tan2021logan,chen2021towards,duan2018weakly,lin2020weakly,song2020weakly,lv2023counterfactual} apply weak-supervised techniques on TSG, where locations of ground truth moments are unavailable during the training stage~\cite{lan2023survey}. However, these specialized methods trained on specific datasets consume high computational resources and suffer from fitting bias~\cite{lan2023survey,yuan2021closer}. To alleviate these problems, many works~\cite{lin2023univtg,liu2022umt,yan2023unloc,lei2021detecting} resort to pretraining with diverse video-language tasks in a more unified framework to boost TSG. However, existing TSG methods mainly focus on short videos, leaving TSG in long videos largely unexplored, which usually involve temporal reasoning under much more complicated contexts.

Meanwhile, recent TSG works introduce more modalities from videos for better localization. A few literatures incorporate modalities, such as optical flows~\cite{anne2017localizing,chen2021end,liu2023exploring} and audio~\cite{chen2020learning,liu2022umt,chen2023curriculum} besides the RGB frames. However, the speech modality has not been comprehensively explored, which sometimes contributes the most to localization, especially in long videos like news, courses, \etc

Therefore, in this paper, we solve TSG in long videos via LLM with both visual and speech modalities integrated.

\spara{Large Language Models~(LLM).}
Recent advancement in Large Language Models~(LLM)~\cite{openai2023gpt4,brown2020language,touvron2023llama,chiang2023vicuna,du2021glm} has made a great difference for its remarkable abilities in Chain-of-Thought~(CoT)~\cite{wei2022chain}, In-Context Learning~(ICL)~\cite{brown2020language}, \etc
CoT decomposes complex tasks into a series of intermediate reasoning steps~\cite{yin2023survey}, while ICL encourages LLM to learn from analogy~\cite{dong2022survey}, empowering LLM with the capability of few-shot learning. For the purpose of handling multimodal tasks with its powerful reasoning ability, Multimodal Large Language Models~(MLLM)~\cite{yin2023survey,gao2023llama,zhu2023minigpt} emerge, which take both multimodal features and texts as inputs.
Some MLLM for videos~(MLLM-V) literature~\cite{chen2023videollm,zhang2023video,li2023videochat,maaz2023video} project video feature sequences into the token embedding space of LLM to have LLM to understand videos. However, they fail to understand the TSG task well and are deficient in temporal perception. In contrast, we take an alternative approach in that we textualize video inputs with temporal marks and activate LLM to conduct TSG with our Boundary-Perceptive Prompting strategy.

\spara{Long Video Understanding.}
Handling long videos imposes substantial demands on computational resources and memories. 
Efficient sparse sampling is often a valid mechanism to better cover long video inputs with controllable memory usage. Such sampling strategies are mainly based on saliency~\cite{korbar2019scsampler,zhi2021mgsampler}, adaptability~\cite{wu2019adaframe,fayyaz2022adaptive}, or stochastics~\cite{rodriguez2023memory}. Since there is much more redundancy in long videos, \cite{barrios2023localizing} removes a great number of irrelevant video segments via multimodal guidance. Since sampling inevitably loses some information, some works~\cite{soldan2022mad,wu2019long} process several video slices and integrate them for global perception.
Besides, information compression improves the perception coverage of videos with condensed representations. \cite{song2023moviechat} designs a memory mechanism to combine short and long memories for long video question answering.
We argue that textualization is also an efficient compression of videos in TSG, especially for the visual modality, thus making it possible to localize moments by LLM.

\section{Proposed Method: Grounding-Prompter}
\label{Method}

We elaborate on the details of our proposed Grounding-Prompter method in this section~(Figure~\ref{fig:framework}). After formulating this problem~(Section~\ref{formulation}), we first transform the TSG task and its multimodal inputs into compressed representations to feed LLM~(Section~\ref{textualization}). Then, to further activate LLM in temporal reasoning, the Boundary-Perceptive Prompting strategy is designed~(Section~\ref{prompting}), where we feed LLM with additional Multiscale Denoising Chain-of-Thought~(CoT)~(Section~\ref{cot}), Validity Principles~(Section~\ref{principles}), and One-Shot In-Context-Learning~(ICL)~(Section~\ref{icl}). With the above, the LLM gives the prediction via step-by-step reasoning.

\subsection{Problem Formulation}
\label{formulation}
Given a video $V$ and a query $Q$, the method is required to predict the start-and-end timestamps of video moments $(\hat{t}_s,\hat{t}_e)$ (second) that matches query $Q$. For LLM-based solutions, the video $V$ with $N$ frames is inputted as textualized representations, speech transcriptions $\{(t_{s(s)}^{(i)},t_{s(e)}^{(i)},s^{(i)})\}_{i=1}^{N_{s}}$ and visual captions $\{(t_{v}^{(i)},c^{(i)})\}_{i=1}^{N_c}$, where $N_s$, $N_c$ are the number of transcriptions and captions, respectively; the $i$-th piece of transcription $s^{(i)}$ lies from time $t_{s(s)}^{(i)}$ to $t_{s(e)}^{(i)}$ and the $i$-th piece of caption is generated on the frame sampled at time $t_{v}^{(i)}$. Usually, $N_s, N_c$ are supposed much smaller than $N$.

\subsection{Compressed Task Textualization}
\label{textualization}
To transform the TSG task and its multimodal inputs into texts that LLM can take in, we design the compressed task textualization pipeline.

To align the behaviors of LLM with the TSG task, we explain the meaning of TSG and formulate the format of its inputs and outputs to LLM, shown in Section~\ref{sec:prompt}(1).
Then, to have LLM understand multimodal inputs, shown in Section~\ref{sec:prompt}(2-4), we textualize speeches and visual modalities into transcriptions and captions with temporal marks. We argue that this textualization preserves enough semantics for localization, thereby an efficiently compressed representation of long videos, which is verified in our experiments~(Section~\ref{experiments}).

Speeches are transcribed into non-overlapped sentences that are temporally partitioned via Automatic Speech Recognition~(ASR).
Since huge amounts of frames in long videos contain much redundancy, a straightforward but effective sampling strategy is adopted, where we pre-detect scene transformations and only sample frames in alignment with scenes and speeches. The intermediate frame for each piece of transcriptions and each scene is selected, \ie the frame at time ${(t_{s(s)}^{(i)}+t_{s(e)}^{(i)})}/{2}$ for $i$-th piece of transcriptions and the same for each scene. This is mainly based on the prior observation that visual contents are usually positively correlated with the synchronized speeches and similar scenes usually share largely redundant semantics. In order to avoid excessive sampling sparsity, we set the minimum number $\overline{N_c}$ of the sampled frames and uniformly sample from the videos up to $\overline{N_c}$ frames when $N_c < \overline{N_c}$. Captions are only generated on these sampled frames. After that, we feed transcriptions and captions with the query text to LLM as the task inputs.

We note that it's possible that the visual captions could be quite noisy, however, the information of the captions is still able to benefit moment localization, which is proved in our experiments.

\subsection{Boundary-Perceptive Prompting}
\label{prompting}

Since mainstream generation tasks in LLM hardly consider temporal boundary perception under complicated long contexts, it's difficult for LLM to accurately follow the TSG task. Additionally, LLM outputs free-form responses, making it possible to generate unreasonable predictions or even fail to provide predictions. To handle them, we propose the Boundary-Perceptive Prompting strategy. First, we design the Multiscale Denoising Chain-of-Thought to enhance temporal boundary perception by reasoning step-by-step under long and noisy contexts. Moreover, several validity principles are introduced to regularize the answer of LLM. Besides, we integrate one-shot ICL as LLMs are few-shot learners~\cite {brown2020language}.

\subsubsection{Multiscale Denoising Chain-of-Thought}
\label{cot}
We decompose our Multiscale Denoising CoT design into the following steps to combine global and local semantics with noise resistance when generating predictions.

\spara{Step 1: Global Understanding.}
We have LLM to summarize the whole video, just shown in Section~\ref{sec:prompt}(5), to filter detailed redundancy and unnecessary repetition with global high-level semantics preserved.

\spara{Step 2: Noise Evaluation.}
Since visual captions usually contain larger amounts of noise compared to speech transcriptions, thus, as illustrated in Section~\ref{sec:prompt}(6), we prompt LLM this observation and instruct LLM to evaluate how the captions assist in moment localization and balance visual-speech information gaps adaptively.

\spara{Step 3: Partition Understanding.}
To encourage LLM to understand local details in both matched and mismatched moments, as shown in Section~\ref{sec:prompt}(7), we have LLM to partition the video by the yet-to-be-predicted start-and-end timestamps and summarize each partition with the given query as conditions, to inspire LLM to capture differences relevant to query among several parts.

\spara{Step 4: Prediction.}
Finally, we have LLM to predict the timestamps $(\hat{t}_s,\hat{t}_e)$ according to the reasoning steps above.

\subsubsection{Validity Principles}
\label{principles}
We set up the following three validity principles to constrain the behaviors of LLM and ensure that a reasonable prediction can be obtained.

\spara{Format Compliance.}
To better improve the LLM's compliance with our multiscale denoising CoT, we devise an answering template in JSON, whose keys cover the reasoning steps in Section~\ref{cot}, demonstrated in Section~\ref{sec:prompt}(8). With format compliance principles, the answers of LLM would be parsed automatically. 

\spara{Answer Regularization.}
To avoid improper predictions, as Section~\ref{sec:prompt}(9) shows, we inform LLM to regularize its answer. First, there exists only one appropriate moment given a query. Furthermore, $\hat{t}_s < \hat{t}_e$ must be ensured for each prediction.

\spara{Plagiarism Prohibition.}
To prevent LLM from copying the prediction of the given example, therefore, we emphasize that LLM should imitate the format and reasoning steps from the example, rather than just copying the prediction, referring to Section~\ref{sec:prompt}(10).

\subsubsection{One-Shot In-Context-Learning}
\label{icl}

With just a one-shot example introduced, illustrated in Section~\ref{sec:prompt}(11), it is proven to significantly enhance temporal reasoning and format compliance. It's fixed during inference for convenience.
\begin{table*}[htbp]
\small
  \centering
  \begin{tabular}{@{}lcccccc@{}}
    \toprule
    Dataset & DiDeMo~\cite{anne2017localizing} & Charades-STA~\cite{gao2017tall} & ActivityNet Captions~\cite{krishna2017dense} & TACoS~\cite{gao2017tall} & VidChapters-mini \\
    \midrule
    Ave Duration~(min) & 0.49 & 0.51 & 1.96 & 4.78 & 13.96 \\
    Modalities & V+A & V+A & V+A & V & V+A+S \\
    \bottomrule
  \end{tabular}
  \caption{Statistical Comparison between VidChapters-mini and other TSG benchmark datasets, where V, A, and S denote visual, audio, and speech modalities.}
  \label{tab:dataset_comparison}
\end{table*}
\subsection{Prompt Example}
\label{sec:prompt}
We present a prompt example in this section to better explain the details of our prompt design.

\spara{(1) Task Description \& Formulation:} You can analyze the correlations between a video and query, and locate the video segment that matches the query. You are given: (1) Video title (2) Query (3) Speech transcription, with temporal information in the format of: [START-TIMESTAMP]-[END-TIMESTAMP]:[TRANSCRIPTION] (4) Visual caption, with temporal information in the format of: [TIMESTAMP]:[CAPTION]. You should give the answer in [X, Y] format where X, Y are the start and end timestamps of the matching segment.
    
\spara{(2) Query:} Habit 2: Build other people up
    
\spara{(3) Speech Transcriptions:} 0-7: While watching clips from my last Game of Thrones video...
    
\spara{(4) Visual Captions:} 5: A woman with long blonde hair... 
    
\spara{(5) Global Understanding:} You summarize the video.
    
\spara{(6) Noise Evaluation:} We note that the visual caption might be quite NOISY. Now you comment if the visual captions are helpful enough for localization. You can give up information from captions if you think some of them are wrong.
    
\spara{(7) Partition Understanding:} You analyze the video content before X, between X and Y, and after Y, respectively. After that, you give the answer [X, Y].
    
\spara{(8) Format Compliance:} Please use JSON format of \{``summary":``..."(you summarize the whole video),``comment": ``..."(you evaluate effectiveness of visual captions), ``query":``..."(the query input), ``before X": ``..."(you summarize video before X), ``between X and Y": ``..."(you summarize video between X and Y), ``after Y": ``..."(you summarize video after Y), ``answer": [X, Y]\}.
    
\spara{(9) Answer Regularization:} We ensure there does exist ONE moment matching the query and X is no more than Y.
    
\spara{(10) Plagiarism Prohibition:} You MUST NOT just copy the answer given by the example! X and Y should be replaced by the real start and end timestamps of the moment you find in videos.

\spara{(11) One-Shot In-Context-Learning:} $<$INPUT$>$=$>$ ...Query: Habit 2: Build other people up. Speech transcriptions: 0-7: While watching... Visual captions: 5: A woman... $<$OUTPUT$>$=$>$ \{``summary": ``The video discusses...", ``comment": ``These captions describe a scene where people talk in a show, but provide limited information to understand the video.", ``query": ``Habit 2: Build other people up", ``before 179": ``Talk about...", ``between 179 and 329": ``Talk about...", ``after 329": ``Talk about...", ``answer": [179, 329]\}. Now you solve the following. $<$INPUT$>$=$>$ ... $<$OUTPUT$>$=$>$

\section{Experiments}
\label{experiments}

\subsection{Datasets and Settings}

\textbf{Datasets.}
We adopt the VidChapters-7M~\cite{yang2023vidchapters} dataset for TSG in long videos, which contains 817K open-domain YouTube videos with 7M user-annotated chapters, featuring longer duration and richer modalities with speeches compared with other TSG benchmark datasets. 
Considering the huge amount of videos with various durations in VidChapters-7M, we randomly select 3 chapter annotations for each video whose duration is 13-15 minutes in its test split, resulting in 1830 query-moment pairs in the final test, called VidChapters-mini. Statistical comparison is shown in Table~\ref{tab:dataset_comparison}.

\spara{Evaluation Metrics.}
We follow the metrics ``r@\{$m$\}"(\%), ``mIoU"(\%) and ``r@\{$n$\}s"(\%) that are widely adopted in TSG. r@\{$m$\} is defined as the percentage of predictions whose Intersection-over-Union~(IoU) with the ground truth is larger than threshold $m$, while r@\{$n$\}s as the percentage of predictions for which the start-time falls within $n$ seconds from the corresponding ground truth start-time. To measure the instruction-following ability of LLM in TSG, a new metric, collapse rate ``cr"(\%), is introduced, which is defined as the proportion of predictions failing to generate answers adhering to the prescribed format. If a collapse happens, the model is regarded as giving an invalid answer, with its IoU being 0.0 and the distance to the ground truth start-time being +inf.

\spara{Implementation Details.}
We apply GPT-3.5-turbo-16k~\footnote{\url{https://platform.openai.com/docs/models/gpt-3-5}} as the LLM model. The video scenes are detected by PySceneDetect~\footnote{\url{https://www.scenedetect.com/}} tool based on content transition, and the sampled frames are captioned by BLIP~\cite{li2022blip} model. For speech transcriptions, we take the Whisper-based~\cite{radford2023robust} tools to finish ASR following~\cite{yang2023vidchapters}. The temperature of LLM is set to 0.0 in all experiments for reproducibility and we set the minimum number of sampled frames $\overline{N_c}$ to 100.

\begin{table*}[htbp]
\small
  \centering
  \begin{tabular}{@{}lrrrrrrrrrrrr@{}}
    \toprule
    Methods & Modalities & r@0.3 & r@0.5 & r@0.7 & r@0.9 & mIoU & r@1s & r@3s & r@5s & r@10s & cr \\
    \midrule
    Random  & - & {13.10} & 5.36 & 1.67 & 0.19 & 10.31 & 0.37 & 0.83 & 1.38 & 2.52 & - \\
    Complete & - & 12.62 & 3.77 & 1.04 & 0.16 & {15.94} & {10.11} & {10.16} & {10.27} & {11.04} & - \\
    CLIP Zero-shot & V & 12.08 & {6.01} & {2.13} & {0.60} & 10.15 & {1.09} & {3.17} & {5.08} & {7.65} & - \\
    BERT Zero-shot & S  & {15.25} & {6.94} & {2.95} & {0.49} & {12.80} & 0.77 & 2.13 & 3.17 & 4.59 & - \\
    \midrule
    VideoChat & V+S & {1.20} & {0.38} & {0.05} & 0.00 & {1.48} & {3.61} & {4.27} & {4.76} & {6.24} & \underline{43.84}\\
    Video-ChatGPT & V+S & {1.20} & {0.60} & {0.05} & {0.05} & {1.21} & 2.57 & 3.01 & 3.55 & 4.32 & {46.72} \\
    Video-LLaMA & A+V+S & {0.60} & 0.24 & {0.12} & {0.06} & 1.16 & {2.60} & {3.50} & {3.98} & {5.25} & 52.81\\
    \midrule
    M-DETR Zero-shot & V & 14.59 & 6.34 & 2.30 & 0.33 & 11.68 & 1.15 & 2.51 & 3.28 & 5.08 & - \\
    M-DETR & V & \textbf{49.48} & \textbf{36.85} & \textbf{26.41} & \textbf{10.93} & \underline{25.41} & \underline{10.33} & \underline{15.53} & \underline{18.97} & \underline{27.99} & - \\
    \midrule
    \textbf{Ours} & V+S & \underline{34.81} & \underline{22.95} & \underline{14.92} & \underline{6.28} & \textbf{26.81} & \textbf{17.60} & \textbf{25.41} & \textbf{32.02} & \textbf{39.73} & \textbf{10.27} \\
    \bottomrule
  \end{tabular}
  \caption{Comparison with other technical roadmaps on VidChapters-mini, where Moment-DETR~(M-DETR) Zero-shot is pretrained with visuals. A, V, and S are short for audio, visual, and speech modalities, which are the input modalities from videos. The best and second are highlighted by \textbf{bold} and \underline{underline}.}
  \label{tab:overall_performance}
\end{table*}

\begin{figure*}[htbp]
    \centering
    \begin{subfigure}[b]{0.32\linewidth}
        \includegraphics[width=1.05\textwidth]{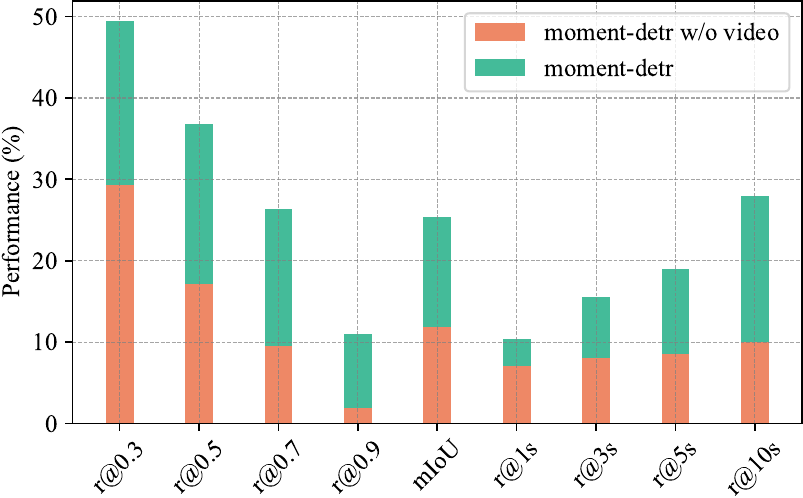}
        \caption{}
        \label{fig:m-detr-1}
    \end{subfigure}
    \hfill
    \begin{subfigure}[b]{0.32\linewidth}
        \includegraphics[width=1.05\textwidth]{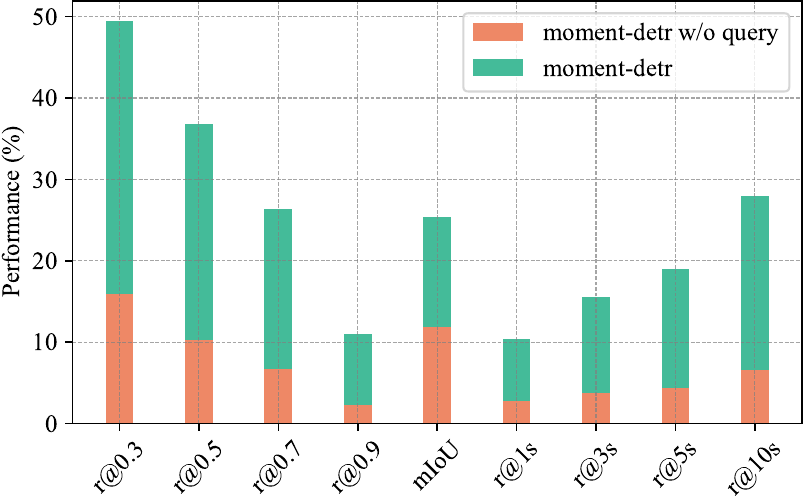}
        \caption{}
        \label{fig:m-detr-2}
    \end{subfigure}
    \hfill
    \begin{subfigure}[b]{0.32\linewidth}
        \includegraphics[width=1.05\textwidth]{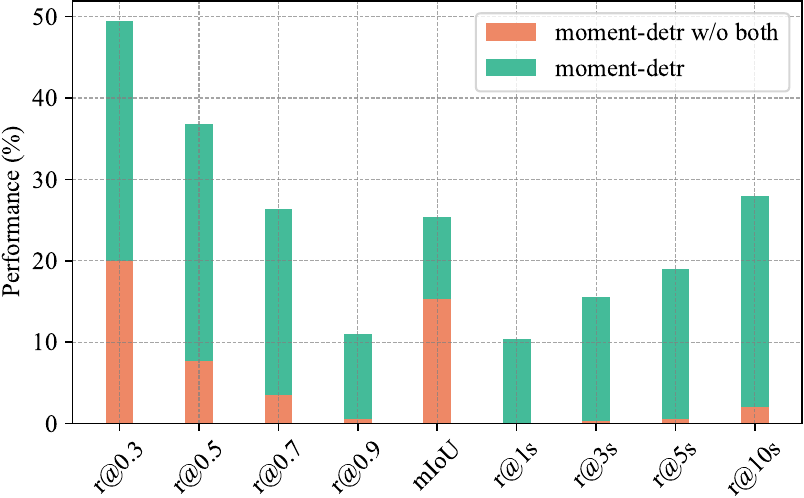}
        \caption{}
        \label{fig:m-detr-3}
    \end{subfigure}
    \caption{Comparisons between moment-DETR and its variants, where the green and orange bars are overlapped in the bottom to clearly illustrate the performance gaps. Moment-DETR still demonstrates comparable performance even some inputs are removed, showing that it largely suffers from fitting bias.}
    \label{fig:m-detr}
\end{figure*}

\subsection{Technical Roadmap Baselines}

We compare our method with three types of technical baselines: the rule-based, Multimodal Large Language Models for Videos~(MLLM-V), and state-of-the-art TSG models.

\spara{The Rule-Based.} We implement four ones.
(i) {``Random"}: We randomly select a pair $(\hat{t}_s,\hat{t}_e)$ with $\hat{t}_s < \hat{t}_e$, and the average metrics are reported after 10 repetitions. 
(ii) {``Complete"}: We select the pair of $(0, T)$ where $T$ is the video duration.
(iii) {``CLIP Zero-shot"}~\cite{radford2021learning}: we calculate the features for uniformly sampled frames and the query via CLIP. Subsequently, the pair $(\hat{t}_s,\hat{t}_e)$  is determined where the highest matching score is attained at time $\hat{t}_s$, and it first decreases by $\epsilon$ at time $\hat{t}_e$, where $\epsilon$ is a hyperparameter.
(iv) {``BERT Zero-shot"}~\cite{kenton2019bert}: similarly, we generate embeddings for each piece of transcriptions and the query. The prediction is obtained through similarities, like CLIP Zero-shot.
For CLIP Zero-shot and BERT Zero-shot, we set $\epsilon$ to 0.05 in accordance with \cite{yang2023vidchapters}.

\spara{MLLM-V.}
We compare with three methods, \ie VideoChat~\cite{li2023videochat}, Video-ChatGPT~\cite{maaz2023video}, and Video-LLaMA~\cite{zhang2023video} in this category. For a fair comparison, these MLLM-Vs take the whole video and its speech transcriptions as input. We sample as many frames as possible from videos to integrate more visual information and also feed them with a similar one-shot example if supported by the MLLM-V. For automatic purposes, format control is adopted as well in order to obtain regularized answers.

\spara{TSG Models}. Furthermore, we also compare our method with the state-of-the-art TSG method, moment-DETR~\cite{lei2021detecting}. We test it in both training-free and training-based fashions. In the training-free fashion, we apply the checkpoint pretrained on QVHighlights~\cite{lei2021detecting} dataset and transfer it to VidChapters-mini. Additionally, in the training-based fashion, a moment-DETR is fully trained on the training split of VidChapters-7M. Additionally, we report three other results of this trained model, by clearing the query, setting visual features to zero, and implementing both of them to check the bias.

\begin{table*}[htbp]
\small
  \centering
  \begin{tabular}{@{}cccrrrrrrrrrr@{}}
    \toprule
    Methods & \ & \ & r@0.3 & r@0.5 & r@0.7 & r@0.9 & mIoU & r@1s & r@3s & r@5s & r@10s & cr \\
    \midrule
    Random & \ & \ & 13.10 & 5.36 & 1.67 & 0.19 & 10.31 & 0.37 & 0.83 & 1.38 & 2.52 & - \\
    Complete & \ & \ & 12.62 & 3.77 & 1.04 & 0.16 & 15.94 & 10.11 & 10.16 & 10.27 & 11.04 & - \\
    \midrule
    \multirow{8}{*}{\textbf{Ours}} & CoT & ICL\\
    \ & \ding{56} & \ding{56} & 2.40 & 1.26 & 0.38 & 0.05 & 4.37 & 12.84 & 18.85 & 23.50 & 29.89 & \textbf{0.00} \\
    \ & \checkmark & \ding{56} & 12.73 & 6.99 & 3.33 & 0.98 & 11.56 & 16.39 & 23.61 & \underline{29.62} & \underline{37.27} & \underline{3.66} \\
    \ & \ding{56} & \checkmark & 21.69 & 13.06 & 7.32 & 2.24 & 17.24 & 16.39 & 23.44 & 27.6 & 34.32 & \textbf{0.00} \\
    \cmidrule(lr){2-13}
    \ & Speech & Visual\\
    \ & \checkmark & \ding{56}  & \underline{28.25} & \underline{18.42} & \underline{11.37} & \underline{4.70} & \underline{21.84} & \underline{16.78} & \underline{23.93} & 29.34 & 36.12 & 18.03 \\
    \ & \ding{56} & \checkmark & 20.22 & 10.49 & 5.14 & 1.15 & 15.17 & 5.90 & 10.27 & 12.40 & 15.68 & 5.08 \\
    \cmidrule(lr){2-13}
    \ & \checkmark & \checkmark   & \textbf{34.81} & \textbf{22.95} & \textbf{14.92} & \textbf{6.28} & \textbf{26.81} & \textbf{17.60} & \textbf{25.41} & \textbf{32.02} & \textbf{39.73} & 10.27 \\
    \bottomrule
  \end{tabular}
  \caption{Ablation Studies on VidChapters-mini. The best and the second are highlighted by \textbf{bold} and \underline{underline}, respectively.}
  \label{tab:ablations}
\end{table*}

\subsection{Empirical Results}
Table~\ref{tab:overall_performance} presents the performance on VidChapters-mini, which indicates that our method achieves the state-of-the-art in training-free settings. We could even beat the fully-trained moment-DETR on more than half of the metrics, especially on r@\{$n$\}s.

Rows 5-7 show the MLLM-Vs completely collapse when handling TSG. They fail to provide answers that follow specific formats on around half of the samples and are even beaten by ``random" and ``complete" on IoU-based metrics, indicating the drawbacks of these MLLM-Vs to understand the TSG task and perceive temporal boundaries across multiple modalities.

Observing rows 8-12, though moment-DETR achieves the highest performance on most IoU-based metrics, however, we observe that the ``state-of-the-art" performance largely derives from fitting distribution bias of both queries and video moments, indicating that it mainly learns spurious correlations other than real semantics, as shown in Figure~\ref{fig:m-detr}. This may explain the reason why moment-DETR only shows marginal improvements compared to CLIP and BERT in zero-shot settings. In contrast, without training, our method shows a high level of generalization capability.

\subsection{Ablation Analysis}

We conduct the following ablation studies in Table \ref{tab:ablations} to evaluate the crucial components in our method. For rows 3-5, we remove the three steps of CoT in Section~\ref{cot} and prompt LLM to merely generate a pair of timestamp predictions in column w/o CoT, and instruct LLM to give answers without one-shot example in column w/o ICL. For rows 6-7, we remove the speech transcriptions or visual captions in both the one-shot example and the test input, to evaluate the effectiveness of multimodal information. 

\subsubsection{Effect of Boundary-Perceptive Prompting}

For rows 3-5, our findings indicate the effectiveness of our Boundary-Perception Prompting strategy, with around 40- and 125-fold improvements on tighter r@0.7 and r@0.9 metrics when both CoT and ICL are implemented. In addition, there are more in-depth observations.
\begin{itemize}
    \item CoT and ICL mainly boost performance by improving temporal coverage of predictions, \ie better predicting the start and end timestamps jointly other than improving the accuracy of point localization. LLM is capable of predicting a single timestamp without careful prompt designs on r@\{$n$\}s but it's beaten even by ``random" and ``complete" on r@\{$m$\} and mIoU without CoT and ICL~(row 3), indicating both CoT and ICL are indispensable for LLM to truly understand the TSG task.
    \item Compared to learning from CoT, LLM is better at imitating examples to understand this novel and complicated temporal task. The implementation of ICL yields performance on r@\{$m$\} metrics that is around \textit{twice} as much as that of CoT, though they achieve comparable performance on r@\{$n$\}s metrics~(rows 4-5).
    \item More reasoning steps activate LLM to understand the novel TSG task but hinder format compliance. One-shot ICL activates imitating ability well with significant performance advancement, but it seems anti-intuitive that there are more format collapses when it's implemented. Interestingly, it has been observed that LLM usually conveys meanings such as ``The answer cannot be determined due to the absence of any existing matches" when format collapses occur, showing interpretation for failure cases to some extent.
\end{itemize}

\subsubsection{Effect of Multimodal Information}

Rows 6-8 show our method comprehensively combines speech and visual modalities, outperforming itself with great margins when any of them is removed. (i) Speeches make more contributions to localization with a 5.13\% improvement on r@0.9, and performance also improves significantly when captions are added, with a 1.58\% update on r@0.9. (ii) The start point prediction accuracy is close to the optimal only with ASR, but captions work well in improving the IoU accuracy. (iii) Although some key information is lost and much noise is introduced when translating videos into sparse captions, LLM still benefits from visual information, showing its strong ability to resist noise. (iv) Speeches contribute to more performance improvement but also cause a higher likelihood of collapse occurrences. The collapses are mitigated when visual captions are integrated, indicating that integrating multimodal information is advantageous for the robustness of our method.

\subsection{Qualitative Analysis}

Compared to traditional TSG methods, our method not only provides structural predictions, but more explanatory remarks as well. In this section, we conduct qualitative analysis on GPT-3.5-turbo-16k.

\begin{figure*}[htbp]
    \centering
    \begin{subfigure}[b]{0.33\linewidth}
        \includegraphics[width=\textwidth]{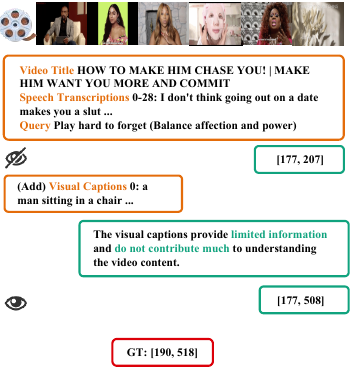}
        \caption{}
        \label{fig:case_study_mm1}
    \end{subfigure}
    \hfill
    \begin{subfigure}[b]{0.33\linewidth}
        \includegraphics[width=\textwidth]{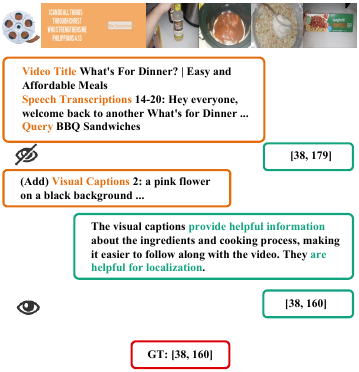}
        \caption{}
        \label{fig:case_study_mm2}
    \end{subfigure}
    \hfill
    \begin{subfigure}[b]{0.33\linewidth}
        \includegraphics[width=\textwidth]{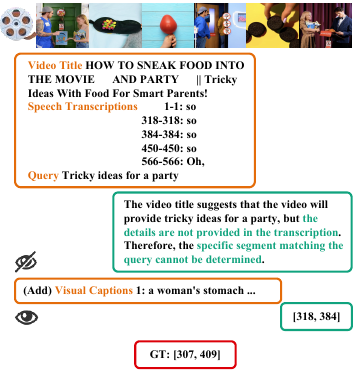}
        \caption{}
        \label{fig:case_study_mm3}
    \end{subfigure}
    \caption{Case demonstration on the multimodal settings. For clarity and brevity, we present them in multi-round dialogs to show how we vary the input we feed the LLM, but in fact, our method only requires one round interaction with LLM. we only demonstrate relevant words with multimodal from LLM's responses. \textcolor[rgb]{0.894,0.424,0.039}{Orange} inputs, \textcolor[rgb]{0.063,0.639,0.494}{green} outputs.}
    \label{fig:case_study_multimodal}
\end{figure*}

\begin{figure*}[htbp]
    \centering
    \begin{subfigure}[b]{0.33\linewidth}
        \includegraphics[width=\textwidth]{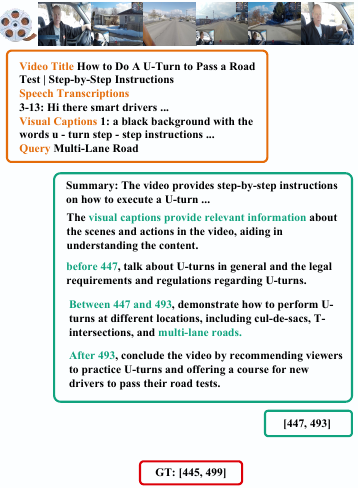}
        \caption{}
        \label{fig:case_study_prompt_1}
    \end{subfigure}
    \hfill
    \begin{subfigure}[b]{0.33\linewidth}
        \includegraphics[width=\textwidth]{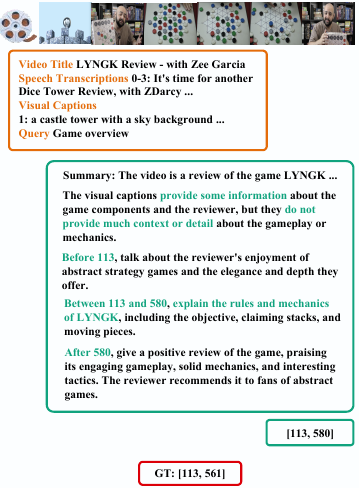}
        \caption{}
        \label{fig:case_study_prompt_2}
    \end{subfigure}
    \hfill
    \begin{subfigure}[b]{0.33\linewidth}
        \includegraphics[width=\textwidth]{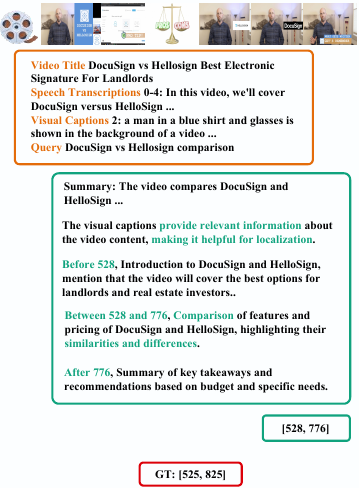}
        \caption{}
        \label{fig:case_study_prompt_3}
    \end{subfigure}
    \caption{Case demonstration on the prompting strategy, where LLM correctly captures the boundaries. \textcolor[rgb]{0.894,0.424,0.039}{Orange} inputs, \textcolor[rgb]{0.063,0.639,0.494}{green} outputs.}
    \label{fig:case_study_prompt}
\end{figure*}

Figure~\ref{fig:case_study_multimodal} demonstrates how our method benefits from multimodal information. In Figure~\ref{fig:case_study_multimodal}(a), we find our method able to resist noise. LLM properly evaluates the noise in visual captions and filters noisy information in captions adaptively, thereby giving a more precise prediction.
In Figure~\ref{fig:case_study_multimodal}(b), it's observed that our method can refine predictions through valid visual captions. LLM concludes that visual captions are beneficial to localization and it refines the end-time prediction. In Figure~\ref{fig:case_study_multimodal}(c), the LLM fails to provide answers solely based on speeches. However, a reasonable prediction is obtained when incorporating visual captions.

To demonstrate the temporal reasoning ability of our method with the Boundary-Perceptive Prompting strategy, a few examples are exhibited in Figure~\ref{fig:case_study_prompt}, showing that our method provides appropriate temporal partitions and correct moment summaries correspondingly.

\section{Conclusion}

In this paper, we propose a Grounding-Prompter method to solve the Temporal Sentence Grounding (TSG) task in long videos.
With the compressed task textualization, we effectively activate LLM to understand the TSG task and its multimodal inputs, speeches and visual content. After that, we propose the Boundary-Perceptive Prompting strategy, which is proven to enhance temporal reasoning and boundary perception under complicated and noisy long contexts.
Experiments prove that our proposed Grounding-Prompter can effectively generate answers with consistent format compliance and achieve state-of-the-art performance with great margins compared to other baseline methods.

To the best of our knowledge, we are the first to explore LLMs for TSG in long videos by reformulating TSG into a long-textual task. With novel prompt designs, the competitive performance of our method indicates the potential of LLM to conduct temporal video tasks in a novel training-free manner. It will be interesting to investigate tuning tailored LLMs for TSG related tasks in future research. 

{
    \small
    \bibliographystyle{ieeenat_fullname}
    \bibliography{main}
}


\end{document}